\documentclass{article}

\usepackage{arxiv}

\usepackage[utf8]{inputenc}
\usepackage[T1]{fontenc}
\usepackage{hyperref}
\usepackage{url}
\usepackage{booktabs}
\usepackage{amsfonts}
\usepackage{amsmath}
\usepackage{amssymb}
\usepackage{natbib}
\usepackage{nicefrac}
\usepackage{microtype}
\usepackage{graphicx}
\usepackage{pdflscape}
\usepackage{longtable}
\usepackage{placeins}

\title{VendorBench-100: A Unified Cross-Paradigm Benchmark for Deepfake Image Detection}

\author{
Sharayu N. Deshmukh \\
Universidade da Beira Interior \\
Covilha, Portugal \\
\texttt{d.sharyu.nilesh@ubi.pt} \\
\And
Md Rashidunnabi \\
Universidade da Beira Interior \\
Covilha, Portugal \\
\texttt{md.rashidunnabi@ubi.pt} \\
\And
Nelton Tiago Gemo \\
Universidade da Beira Interior \\
Covilha, Portugal \\
\texttt{nelton.gemo@ubi.pt} \\
\And
Kurundkar G. D.\\
Department of Computer Science \\
Shri Guru Buddhi Swami College \\
Purna, India \\
\texttt{gajanan.kurundkar@gmail.com} \\
\And
Mahamune M. R. \\
School of Computational Science \\
Swami Ramanand Tirtha Marathwada University \\
Nanded, India \\
\texttt{mohnish.mahamune@gmail.com} \\
\And
Nilesh K. Deshmukh \\
School of Computational Science \\
Swami Ramanand Tirtha Marathwada University \\
Nanded, India \\
\texttt{nileshkd.srt@gmail.com} \\
}

\begin{document}
\maketitle

\begin{abstract}
\noindent Deepfake image detection is served by three fundamentally different paradigms that is commercial APIs, zero-shot vision-language models (LLMs), and open-source detectors that, despite their widespread use, are rarely evaluated under a common protocol, making direct comparison difficult. We introduce \textit{VendorBench-100}, a cross-paradigm benchmark that evaluates 36 representative models using a single adversarial 100-image corpus, a unified output schema, and a common evaluation framework. To ensure reliable assessment under the corpus's intentional class imbalance, models are ranked primarily by the Matthews correlation coefficient (MCC), with ROC-AUC reported as a threshold-independent measure of ranking ability.
Rather than maximizing size, it emphasizes real-world difficulty through a taxonomy of eight edge-case families such as face swaps, text-to-video stills, AI photo edits, avatar compositing, opaque-provenance images, and compressed research frames.
Commercial APIs achieve the strongest median performance, followed by vision LLMs and open-source detectors, though individual open-source models remain competitive with the best LLMs. Across all 36 models, MCC and ROC-AUC are strongly correlated (Pearson $r \approx 0.86$), so the two metrics mostly agree; the more consequential finding is narrower and one-directional: a subset of otherwise strong rankers are miscalibrated at their shipped default threshold, so a high ROC-AUC can overstate real-world deployability. Separately, and just as importantly, raw accuracy and $F_1$ are unreliable on this corpus's imbalanced class split, since a model that predicts ``fake'' indiscriminately scores deceptively well on both while offering no real discriminative skill. Together, these findings argue that no single metric is safe in isolation: MCC and specificity should always accompany ROC-AUC and accuracy. We release the complete evaluation framework and results to support reproducible research. The source code and data are available at:  \url{https://github.com/sharayu-20/vendorbench-100}
 
\vspace{6pt}

\end{abstract}

\section{Introduction}
 
The past several years have seen an extraordinary acceleration in the quality and accessibility of synthetic media. What once required specialized expertise and hand-tuned generative adversarial networks can now be accomplished by an ordinary user with a consumer-facing web application, a short text prompt, or a single uploaded photograph. Diffusion-based generators produce photorealistic faces and scenes frequently indistinguishable from genuine photographs under casual inspection, while face-swapping and avatar tools allow a person's likeness to be transplanted into arbitrary video or image content in minutes. This democratization has a well-documented dark side: deepfakes and AI-generated imagery have been implicated in large-scale financial fraud carried out through impersonated video calls \citep{cnn2024hkfraud}, non-consensual intimate imagery \citep{umbach2024nsii}, manufacture of child sexual abuse material \citep{iwf2023aicsam}, and coordinated disinformation \citep{lse2025misinfo}. As convincing fakes have become cheap and fast to produce, the burden of distinguishing authentic from synthetic content has shifted from specialized forensic laboratories to ordinary platforms, institutions, and individuals with neither the expertise nor the tooling to perform that analysis themselves.
 
\begin{figure}[!ht]
    \centering
    \includegraphics[width=\textwidth]{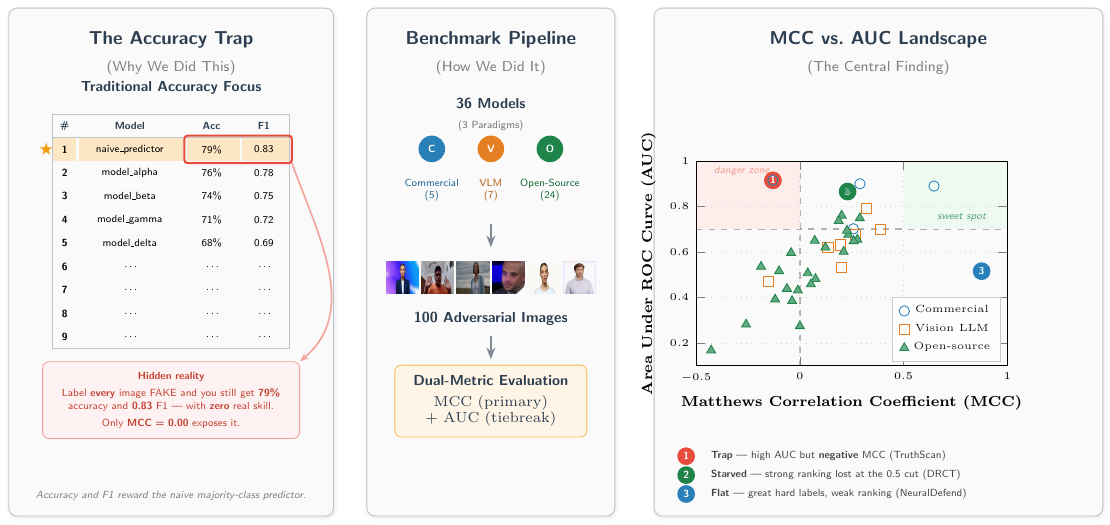}
    \caption{
    \textbf{Overview of VendorBench-100.} Left: the accuracy trap. A naive always-fake predictor scores 79\% accuracy and 0.88 F1 despite an MCC of 0.00. Center: the benchmark pipeline. 36 models across three paradigms scored on 100 adversarial images under a dual-metric protocol (MCC primary, ROC-AUC tiebreak). Right: the MCC-versus-AUC landscape. Most models fall along the diagonal where the two metrics agree (Pearson r $\approx$ 0.86), while a starved subset of otherwise strong rankers sits below it, separating the classes well yet under-performing at their shipped 0.5 threshold.
    }
    \label{fig:teaser}
\end{figure}

Three fundamentally different kinds of tool now compete to fill this gap, and they are almost never compared to one another. Commercial detection APIs let an organization submit an image and receive a verdict without training or hosting a model, and are marketed with prominent accuracy figures that are rarely independently verified. General-purpose vision-language models, never trained specifically for forgery detection \citep{openai2024gpt4,google2024gemini}, are increasingly prompted zero-shot as an ad hoc detector by practitioners who already have API access to them for other purposes. And a large, fragmented ecosystem of open-source detectors, spanning academic checkpoints \citep{chen2024drct,koutlis2024rine}, Hugging Face community classifiers, and purpose-built research models, offers a free alternative to both, at the cost of requiring local GPU infrastructure and offering no vendor accountability. Each of these three paradigms is evaluated, when it is evaluated rigorously at all, in isolation: commercial vendors publish internal benchmarks, vision-LLM providers report general-purpose capability rather than forgery-specific accuracy, and academic detectors are benchmarked against each other on shared research datasets that rarely resemble the messy, adversarial content circulating today \citep{chandra2025deepfakeeval2024,wang2022deepfakesurvey}. A practitioner deciding how to spend a detection budget, pay for a commercial API, prompt a model they already have access to, or self-host a free detector has no single source of evidence that measures all three options the same way. Figure~\ref{fig:teaser} summarizes the paper's motivation and workflow: the accuracy trap that motivates our metric choice (a naive always-fake predictor scores 79\% accuracy yet 0.00 MCC), the benchmark pipeline itself (36 models, three paradigms, dual-metric evaluation), and a preview of the central finding, an MCC-versus-AUC landscape in which most models cluster along the diagonal while a starved subset of otherwise strong rankers sit below it, miscalibrated at their shipped decision threshold.

This paper closes that gap. We construct VendorBench-100, a single, fixed, adversarial corpus of 100 images (79 fake, 21 real) assembled specifically to be hard rather than merely large: every image was hand-picked to probe a distinct failure mode, near-duplicate face swaps, letterboxed video-generation stills, on-device AI photo edits of otherwise-real scenes, and images of deliberately opaque provenance, so that a model's score reflects performance on the difficult, modern middle ground rather than easy separability on saturated public datasets \citep{rossler2019faceforensicspp,zhu2023genimage}. We then evaluate 36 models spanning all three paradigms under one shared task definition, one common output schema, and one abstention policy, and rank them by the Matthews correlation coefficient, with ROC-AUC as a threshold-free tiebreak, because the corpus's 79:21 class skew makes raw accuracy trivially gameable by a detector that simply predicts ``fake'' by default.
 
The result is not a simple ranking in which one paradigm or one model wins outright. Commercial APIs post the strongest typical performance, and the best open-source ranker outperforms every vision LLM in the study on pure ranking power. Across the full leaderboard, a model's ability to separate the classes (ROC-AUC) and its ability to make a well-calibrated decision at its shipped threshold (MCC) are strongly correlated, but a subset of models, disproportionately strong rankers, remain miscalibrated at that threshold, so their ROC-AUC overstates how deployable they actually are. This one-directional gap recurs across all three paradigms and is, we argue, a more actionable finding than any single model's leaderboard position, since it identifies exactly where a headline ranking-power figure can mislead a purchasing or deployment decision. We report these results with an explicit accounting of the corpus's small size and single-run design, and we release the full evaluation harness, per-image evidence, and aggregated results to support independent replication and extension.
Our main contributions are
\begin{itemize}
\item We present VendorBench-100, a cross-paradigm benchmark evaluating 36 models such as 5 commercial APIs, 7 zero-shot vision LLMs, and 24 open-source detectors under one common normalized output schema and one abstention policy, so that results are directly comparable across three paradigms almost never measured on common ground.
\item We construct and document a deliberately hard, edge-case-heavy 100-image corpus organized around eight distinct failure-mode families (live face-swap smearing, near-duplicate online face swaps, letterboxed text-to-video stills, avatar compositing seams, fully synthetic text-to-image output, on-device AI photo edits of real scenes, opaque-provenance manipulations, and compressed research-dataset frames), together with a per-image provenance registry and a filename-based anti-leakage protocol.
\item We show a clear ordering by paradigm as commercial APIs lead, vision LLMs occupy the middle, and most open-source detectors trail, that is nonetheless paradigm-level rather than universal, since individual open-source detectors remain competitive with, or ahead of, the strongest vision LLMs on ranking power \citep{chen2024drct}.
\item We identify a one-directional miscalibration pattern in a subset of otherwise strong rankers, whose ROC-AUC substantially overstates the reliability of their default-threshold decisions (MCC), and we show that this metric-reliability risk, together with the accuracy trap induced by class imbalance, argues for reporting MCC and specificity alongside any headline ranking-power or accuracy figure.
\end{itemize}

\section{Related Work}
\label{sec:related}

This section situates VendorBench-100 within the prior work its three tracks each draw on. We begin with the threat landscape that motivates detection in the first place, then survey the generative techniques and the academic detection methods built to counter them, including the datasets those methods are trained and benchmarked on. We then turn to the two paradigms outside the traditional academic-detector literature that this paper is the first to place alongside it: vision-language models used as zero-shot detectors, and the commercial detection services and provenance-based alternatives already deployed in practice. We close with the still-thin literature on evaluating detectors under real-world rather than laboratory conditions, and the evaluation methodology our own metric choices draw on.

\subsection{Threat Landscape and Motivation}

Deepfakes emerged as a societal concern with early landscape reports documenting rapid growth in synthetic media and its concentration in abusive applications \citep{ajder2019deeptrace,hsh2023state}. High-value financial fraud has since been carried out through real-time video impersonation, in one widely reported case costing a multinational firm tens of millions of dollars \citep{cnn2024hkfraud}; regulators have issued formal warnings to financial institutions about such schemes \citep{fincen2024deepfake}. Beyond financial harm, cross-national survey work has documented the prevalence and psychological impact of non-consensual synthetic intimate imagery \citep{umbach2024nsii}, and child-safety organizations have reported a troubling rise in AI-generated child sexual abuse material \citep{iwf2023aicsam}. Broader assessments extend this threat picture further still, documenting risks to election integrity through synthetic disinformation \citep{lse2025misinfo}, to financial-sector cybersecurity more generally \citep{wef2025gco}, to minors specifically \citep{eprs2025children}, and to the digital identity-verification systems that many online services rely on \citep{wef2026cybercrimeatlas}. Taken together, this body of work establishes both why reliable detection matters and why the tools organizations actually deploy to perform that detection deserve careful, independent measurement rather than uncritical trust in vendor marketing.

\subsection{Generative Models and Detection of Synthetic Imagery}

The images a detector must classify are the product of a diverse, rapidly evolving generative ecosystem. Generative adversarial networks and style-based architectures established photorealistic face synthesis \citep{goodfellow2014gan,karras2019stylegan,karras2020stylegan2}, and denoising diffusion probabilistic models together with latent diffusion have since come to dominate high-quality text-to-image generation \citep{ho2020ddpm,rombach2022ldm}, producing a visually distinct artifact family from their GAN-era predecessors. Orthogonal to whole-image synthesis, convolutional face-swapping and neural-rendering methods enable targeted identity replacement within an otherwise real image or video, rather than generating a scene from scratch \citep{korshunova2017faceswap,thies2016face2face,thies2019neuraltextures,korshunov2018deepfakes}. This three-way split including GAN-era synthesis, diffusion-era synthesis, and localized identity replacement is a central reason a detector that performs well on one manipulation family often degrades sharply on another: each leaves behind a different statistical fingerprint, and a model implicitly or explicitly tuned to one fingerprint has no guarantee of generalizing to the others.

A parallel literature targets fully synthetic, rather than face-manipulated, imagery, and much of it underlies the open-source detectors we benchmark in Section~\ref{sec:results}. Early work showed CNN-generated images carry detectable fingerprints within their generator family \citep{wang2020cvpr}, while frequency-domain analyses showed the upsampling common to many generators leaves spectral artifacts \citep{frank2020icml,durall2020cvpr}. The central open problem is generalization across unseen generators: pre-trained vision-language feature spaces improve cross-generator robustness over narrowly trained detectors \citep{ojha2023cvpr,cozzolino2023clip}, intermediate-encoder-block representations from large pre-trained backbones offer a further such feature space \citep{koutlis2024rine}, reconstruction-based cues targeting diffusion models' encoding process offer a complementary signal \citep{wang2023dire,ricker2024aeroblade}, contrastive reconstruction-based training has been proposed specifically to improve cross-generator generalization for diffusion-generated images \citep{chen2024drct}, and category-common-prompt injection into CLIP has been proposed as a further generalization strategy \citep{tan2024c2pclip}. Benchmarks such as GenImage quantify exactly this generalization gap at scale, showing that detectors trained on one set of generators can lose substantial accuracy when evaluated on held-out ones \citep{zhu2023genimage}. This pattern shows strong in-distribution performance that degrades under distribution shift is precisely what our open-source track exposes at the level of individually deployable, publicly downloadable checkpoints: most of the 24 detectors we evaluate were not trained on the specific generator mix our adversarial corpus contains, and their scores reflect exactly this generalization failure rather than a defect in the corpus.

\subsection{Deepfake Datasets and Detection Benchmarks}

Empirical progress in face-manipulation detection has been driven largely by the availability of large, labeled corpora, whose evolution tracks how the threat has grown in scale and realism. FaceForensics++ standardized manipulated-face detection around a fixed set of manipulation methods and compression levels \citep{rossler2018faceforensics,rossler2019faceforensicspp}. The DeepFake Detection Challenge dataset substantially expanded both scale and actor diversity \citep{dolhansky2019dfdcpreview,dolhansky2020dfdc}, WildDeepfake introduced content collected directly from the internet rather than generated under laboratory conditions \citep{zi2020wilddeepfake}, and Celeb-DF specifically targeted the visual-quality gap between early, artifact-heavy deepfakes and the higher-fidelity forgeries that had begun to circulate in the wild \citep{li2020celebdf}. More recent efforts have pushed further on both diversity and unification: ForgeryNet consolidates many distinct forgery types under a single benchmark \citep{he2021forgerynet}, and DF40 explicitly targets the next generation of forgery techniques as generative models have continued to advance \citep{yan2024df40}. Consolidated frameworks such as DeepfakeBench standardize the training and evaluation protocol itself, showing how strongly reported accuracy depends on preprocessing and cross-dataset transfer choices \citep{yan2023deepfakebench}. Our own corpus draws on FaceForensics++ and DF40 style frames as two of more than twenty provenance sources, but where this prior work benchmarks trainable detector architectures on large, relatively homogeneous corpora, our object of study is a small, deliberately heterogeneous, adversarial set evaluated against closed commercial APIs, general-purpose vision LLMs, and independently distributed open-source checkpoints simultaneously.

\subsection{Vision-Language Models as Zero-Shot Detectors}

A newer and less-studied question is whether general-purpose vision-language models, never trained specifically for forgery detection, can be prompted zero-shot as usable detectors. Contemporary multimodal LLMs from several providers including OpenAI's GPT family with vision input \citep{openai2024gpt4}, Google's Gemini \citep{google2024gemini}, Anthropic's Claude Opus \citep{anthropic2024claudeopus}, Alibaba's Qwen vision-language models \citep{bai2023qwenvl}, Meta's Llama 4 with multimodal variants including the Maverick configuration \citep{meta2025llama4}, NVIDIA's Nemotron vision-language family \citep{nvidia2024nemotron}, and Zhipu AI's GLM family \citep{zhipu2024glm} expose strong general visual reasoning but no forgery-specific training signal, so their reliability as detectors is an open empirical question rather than an assumption we can import from their general-purpose benchmarks. We treat this question directly by including one representative model from each of these seven families as a dedicated evaluation track (Section~\ref{sec:methodology}), prompted with a shared forensic verdict schema rather than a bespoke prompt per model, and scored under exactly the same metrics as the commercial and open-source tracks.

\subsection{Provenance, Watermarking, and Commercial Detection Services}

Passive, after-the-fact detection is not the only proposed defense against synthetic media. Proactive provenance schemes instead try to establish authenticity at the moment of capture or generation. Content Credentials under the Coalition for Content Provenance and Authenticity's (C2PA) specification bind cryptographically signed provenance metadata directly to a media file \citep{c2pa2021spec,c2pa2024spec}, and internet-scale watermarking schemes such as SynthID embed an imperceptible signal directly into generator outputs at the moment of creation \citep{dathathri2024synthidtext,gowal2025synthidimage}. These mechanisms are valuable but fundamentally proactive: they depend on voluntary adoption and provide no protection against the enormous volume of unmarked content already in circulation, or against generators that decline to participate. Passive detectors, of the kind evaluated across all three tracks in this paper, remain the only recourse for that unmarked content.

The commercial detection services in our first track sit within a broader, still-emerging industry landscape characterized by limited public documentation. Reality Defender \citep{realitydefender2024realapi}, Hive AI \citep{hive2024genai}, Sightengine \citep{sightengine2024genai}, TruthScan \citep{truthscan2024imageapi}, and Neural Defend \citep{neuraldefend2024docs} each expose a different native output schema: some return categorical tags, others per-class confidence scores, and others free-text verdicts. This heterogeneity, now multiplied across three entirely different paradigms rather than one, is precisely what necessitates the unified normalization schema described in Section~\ref{sec:methodology}. Other commercial and research detectors not evaluated here, including Sensity AI \citep{sensity2024platform}, Illuminarty \citep{illuminarty2024api}, AI or Not \citep{aiornot2024api}, and Intel's real-time FakeCatcher \citep{intel2022fakecatcher}, illustrate that the commercial landscape this normalization problem applies to is broader still.

\subsection{Evaluations of Detectors in the Wild and Evaluation Methodology}

A growing body of work asks whether laboratory-reported accuracy survives contact with real-world data, and the answer is consistently discouraging. Empirical studies report substantial generalization gaps for image detectors when evaluated across unseen generators and under realistic degradations such as recompression and resizing \citep{li2023generalizable,lu2023assessment}, and a broad survey specifically catalogs the reliability pitfalls that afflict deepfake detection as a field \citep{wang2022deepfakesurvey}. Some pitfalls are structural: dataset construction can leak shortcuts, with post-processing and JPEG artifacts, rather than synthesis cues, driving an apparently high accuracy \citep{grommelt2024fakeorjpeg}, and ``sanity check'' analyses show ostensibly strong detectors can rely on spurious, dataset-specific signals \citep{yan2025sanity}. Most directly related to our own work, several recent efforts have evaluated detection systems, including named commercial offerings, on content actually circulating in the wild rather than on curated laboratory datasets. Deepfake-Eval-2024 evaluates detectors, including commercial products, on recently circulated real-world content and reports accuracy figures markedly lower than vendor marketing claims \citep{chandra2025deepfakeeval2024}; related commercial and production-style evaluations include ARIA \citep{li2024aria}, ``Organic or Diffused'' \citep{ha2024organic}, the Visual Counter Turing Test \citep{vct2}, AI-GenBench \citep{pellegrini2025aigenbench}, and direct human-versus-AI comparisons of detection accuracy \citep{iufereva2025human}. Robustness studies further show well-performing detectors can be defeated by adversarial perturbations \citep{hussain2021adversarial}, underscoring that accuracy on unperturbed content is only a partial picture of real-world reliability. Where these efforts each cover one or two paradigms, our contribution is a controlled, named, cross-paradigm comparison spanning all three on a single fixed corpus under identical normalization.

Finally, our metric choices draw on classical evaluation methodology well established in machine learning but under-applied to commercial and zero-shot deepfake-detection systems specifically. Systems that may decline to produce a verdict are properly analyzed through the reject-option and selective-classification framework \citep{chow1970reject,elyaniv2010selective,geifman2017selective,geifman2019selectivenet,zadrozny2004sampleselection,choe2023counterfactual}, motivating reporting coverage alongside accuracy computed only over decided images. For imbalanced binary classification problems of the kind our real-versus-fake corpus presents, single scalar summaries such as raw accuracy can be actively misleading, and the Matthews correlation coefficient together with careful, separate reporting of class-conditional error rates is recommended in place of accuracy alone \citep{chico2020mcc,brodersen2010balanced,sokolova2009metrics}. We adopt MCC as our primary ranking metric for this reason, and add threshold-free ROC-AUC as a tiebreak precisely because, as we show in Section~\ref{sec:discussion}, the two can diverge for a specific, identifiable subset of models even when they agree closely across most of a leaderboard.

\FloatBarrier
\section{Methodology}
\label{sec:methodology}

\begin{figure*}[!ht]
    \centering
    \includegraphics[width=\textwidth]{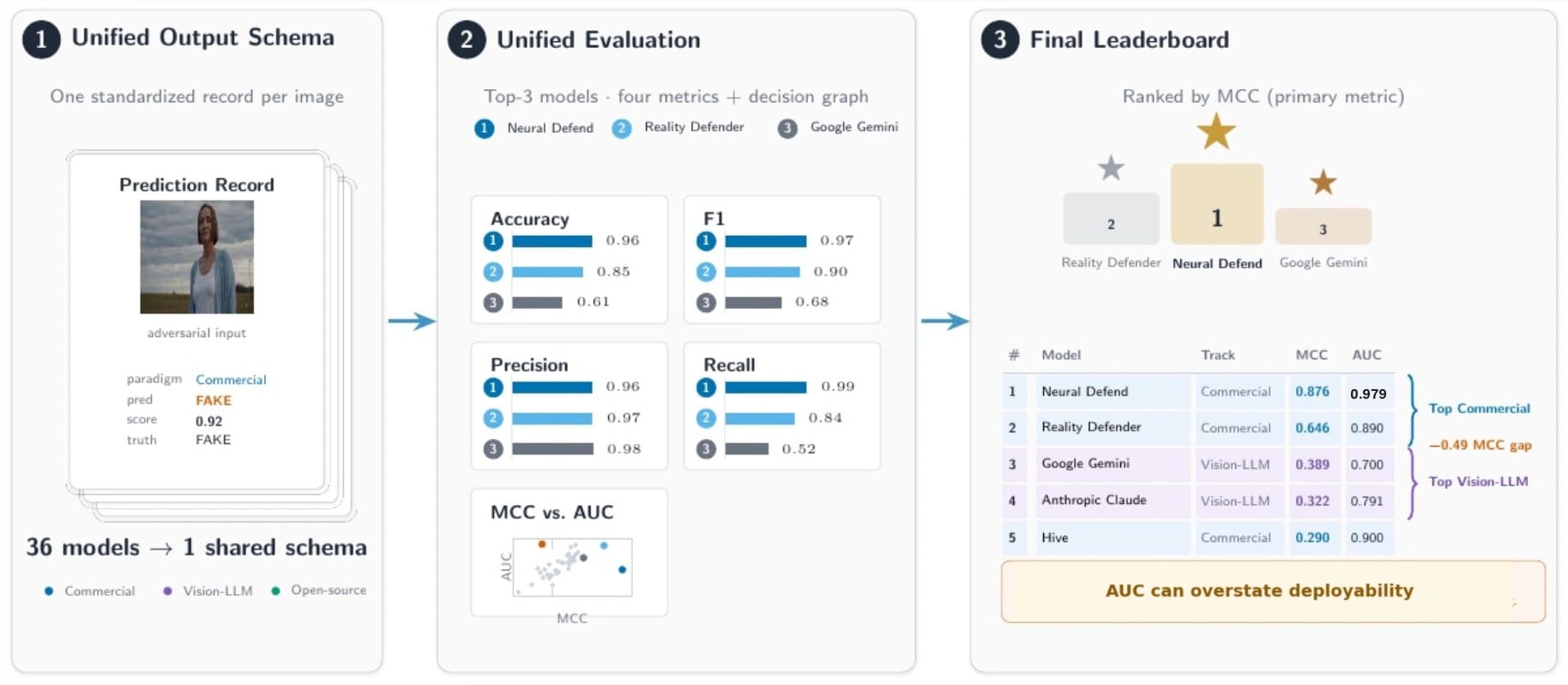}
    \caption{
    End-to-end evaluation architecture of VendorBench-100. Left: every model's output is normalized into one shared prediction record (label, P(fake) $\in$ [0,1], success). Center: a common metrics engine scores all models at the same 0.5 threshold on P(fake), reporting accuracy, precision, recall, specificity, F1, MCC (primary), and ROC-AUC (tiebreak). Right: the MCC-ranked leaderboard, led by Neural Defend on both metrics.
    }
    \label{fig:architecture}
\end{figure*}

Evaluating three fundamentally different detection paradigms fairly requires getting three things right, in order: the corpus every model is measured against, the protocol that makes their heterogeneous outputs comparable at all, and the metrics used to interpret the result. We address each in turn, beginning with the corpus itself, since every downstream comparison inherits whatever strengths and limitations that corpus has. Figure~\ref{fig:architecture} walks through the evaluation architecture: every model is normalized into a shared prediction record, scored by a common metrics engine reporting four core metrics plus the MCC-versus-AUC comparison, and ranked into a single final leaderboard.

\subsection{Dataset}
\label{sec:dataset}

\begin{figure*}[!ht]
    \centering
    \includegraphics[width=\textwidth]{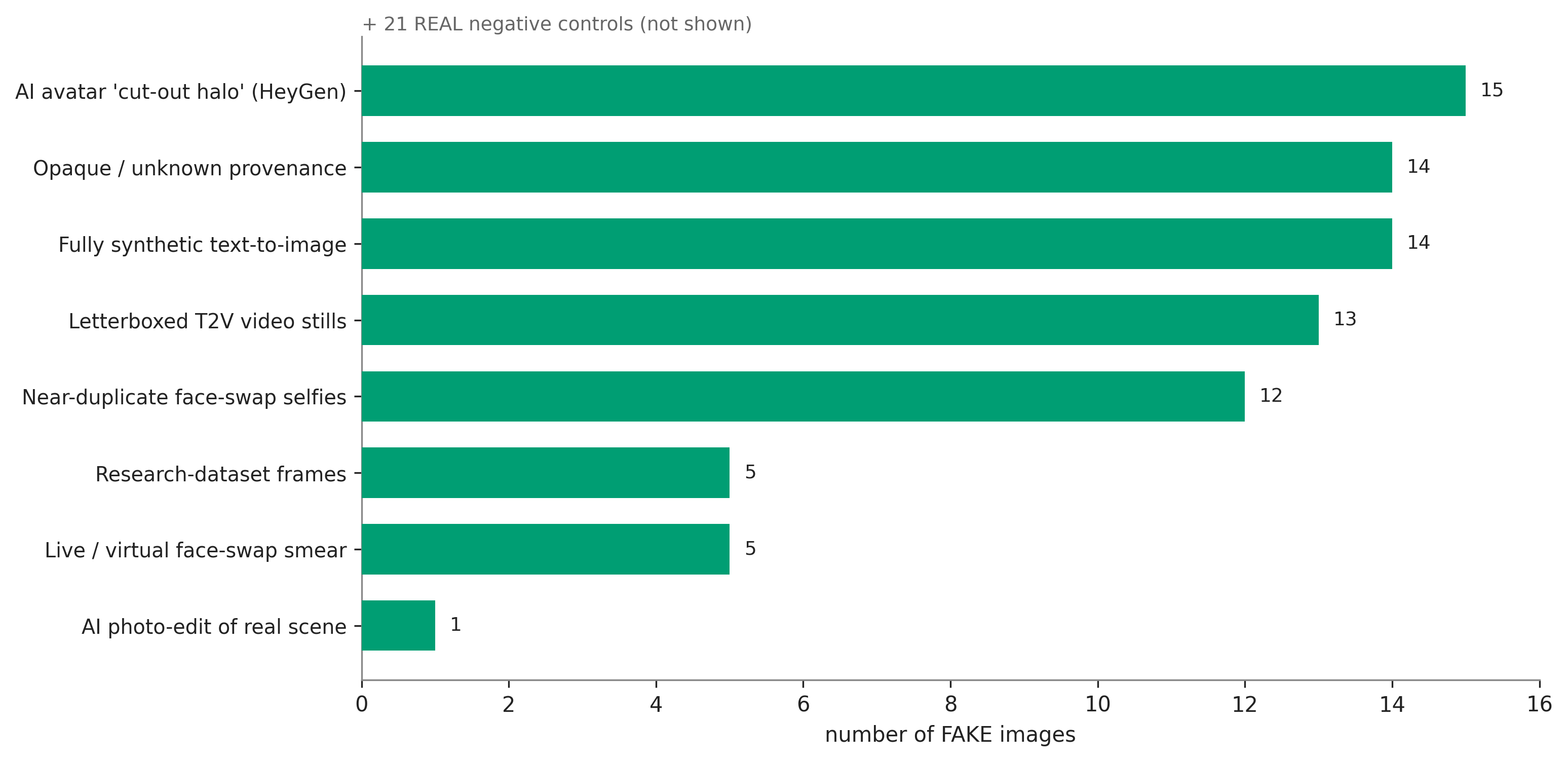}
    \caption{Distribution of the 79 fake images across the eight edge-case families in VendorBench-100. The corpus intentionally emphasizes difficult, real-world failure modes rather than large-scale dataset coverage.}
    \label{fig:taxonomy}
\end{figure*}

We evaluate all 36 models on a single fixed corpus of 100 still images: 79 labeled fake and 21 labeled real (a 79:21 split). Unlike a corpus optimized for scale, this one is optimized for difficulty: every image was hand-selected to probe a distinct, currently under-served failure mode, so that a model's score reflects skill on hard, modern generator output rather than easy separability on saturated public benchmarks. The corpus deliberately spans more than twenty provenance sources so that no single generator, resolution, or lighting regime dominates and no detector can succeed merely by tuning to one artifact profile. Figure~\ref{fig:taxonomy} summarizes the distribution of the 79 fake images across the eight edge-case families included in VendorBench-100. The corpus intentionally emphasizes difficult, contemporary manipulation scenarios rather than balancing category frequencies, ensuring that detector performance reflects robustness under realistic failure modes instead of performance on a homogeneous benchmark.

\subsubsection{Edge-Case Taxonomy}

The fake portion of the corpus is organized around eight distinct edge-case families, each defeating detectors for a structurally different reason. Live face-swap smearing (DeepFaceLive-style capture) entangles manipulation artifacts with ordinary webcam motion blur, so a detector cannot cleanly separate generation artifact from capture noise. Near-duplicate online face swaps (Picsi.Ai, Tuguoba, VidMage, Live3D) replace only the face region on an otherwise genuine selfie, leaving global image statistics looking authentic and punishing models that flip predictions on trivially similar inputs. Letterboxed text-to-video stills (Sora, Veo, Veo 3) are extracted frames carrying black letterbox bars and cinematic color grading that can read as ordinary photographic post-processing rather than a generation artifact. AI avatar compositing seams (HeyGen) exhibit a faint glowing outline around the subject that is a compositing edge, not a pixel-frequency signature, so frequency-domain detectors miss it. Fully synthetic text-to-image output (DALL-E, Gemini, and several additional generators) is the classical generated-image case, though several samples are high-resolution and photorealistic enough that only faint over-smoothness remains as a cue. On-device AI photo edits of otherwise-real scenes (Samsung Galaxy AI) are the single hardest family: because only a small, localized region of an overwhelmingly authentic photograph is synthetic, any whole-image fakeness score is diluted toward ``real.'' Opaque-provenance manipulations carry no independently identifiable generator at all, modeling the realistic ``unknown tool'' scenario a deployed detector actually faces. Finally, compressed research-dataset frames (DF40, FaceForensics++) contribute heavily compressed, low-resolution talking-head content with characteristic waxy skin blending. Compounding all eight families, the corpus skews toward small, heavily recompressed images: 78 of the 100 images are at or below 256 pixels on the long side, which erodes exactly the high-frequency evidence that many frequency-based detectors depend on.

\subsubsection{Composition, Provenance, and Anti-Leakage Protocol}

\begin{figure*}[!ht]
    \centering
    \includegraphics[width=\textwidth]{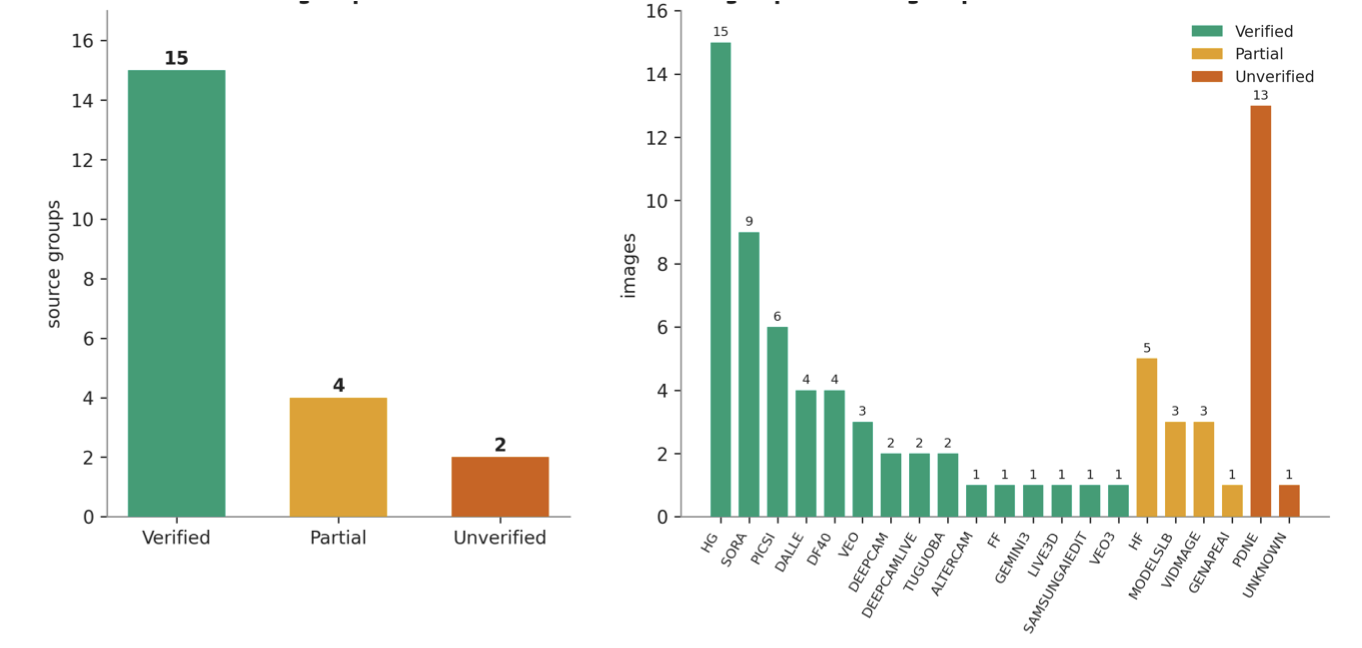}
    \caption{Provenance verification statistics for the fake-image corpus. Left: verification status of the 21 source groups. Right: image counts contributed by each source group.}
    \label{fig:provenance}
\end{figure*}

The 79 fake images are drawn from 21 distinct source groups, verified against public evidence wherever the group's originating platform or dataset is independently identifiable: 15 groups are fully verified (e.g., OpenAI Sora, Google Veo, HeyGen, DF40, FaceForensics++), 4 are partially verified (the platform exists and is named, but the exact model or generation run is not pinned down), and 2 remain deliberately opaque, retained explicitly as unknown-tool cases rather than excluded. The 21 real images are authentic photographs including bare-faced selfies, portraits, and casual indoor and outdoor shots used as negative controls to measure false-positive behavior. A complete per-image provenance registry, including source, scenario, and verification status for every entry, is released alongside the corpus. Figure~\ref{fig:provenance} provides an overview of the corpus provenance. Most source groups are fully verified and account for the majority of images, while a smaller subset consists of partially verified or intentionally opaque sources that emulate realistic unknown-provenance content encountered in deployment.

Because two of the three tracks in this study are systems capable of reading filenames or metadata (vision LLMs in particular), the corpus enforces a strict anti-leakage protocol: every model is served images under neutral, numeric filenames (\texttt{001.jpg}, \texttt{002.jpg}, \ldots) with no label-revealing information in the filename, path, or accompanying metadata. The descriptive, label-bearing identifiers used in the provenance registry (e.g.\ \texttt{fake\_SORA\_003}) are post-hoc bookkeeping keys only and are never presented to a model at evaluation time. Ground truth is held in a separate manifest file, so a correct prediction can only come from the image content itself.

We treat this corpus explicitly as a diagnostic stress test rather than a population-accuracy benchmark: with only 100 images spread across 21 fake source groups, per-group sample sizes are small, and the corpus is far too small to support training, fine-tuning, or a claim about calibrated field accuracy. Its purpose is to surface where and how each of the 36 models breaks under difficult, heterogeneous, adversarial conditions, and we return to this scope limitation explicitly in Section~\ref{sec:limitations}.

\subsection{Three-Track Evaluation Architecture}
\label{sec:normalization}

Evaluating 36 models spanning three fundamentally different deployment paradigms requires three fundamentally different integration strategies, unified by a single canonical output schema. Every model, regardless of track, is normalized into an identical result record consisting of a hard label (\texttt{FAKE} or \texttt{REAL}), a confidence score interpreted uniformly as $P(\text{fake}) \in [0,1]$, and a success flag distinguishing a genuine verdict from an abstention or failed call. This shared schema, not any single track's native format, is what makes commercial, LLM, and open-source outputs directly comparable.

\smallskip
\noindent\textbf{Commercial APIs (5 models).} Each of the five commercial vendors, Neural Defend, Reality Defender, Sightengine, TruthScan, and Hive AI, is queried live over HTTP for every image in the corpus, and round-trip latency is measured directly. This is the only track for which latency is reported, since it is the only track measuring a genuine, comparable, hosted-service round trip.

\smallskip
\noindent\textbf{Vision LLMs (7 models).} Seven general-purpose vision-language models such as Gemini \citep{google2024gemini}, Claude Opus 4.8 \citep{anthropic2024claudeopus}, Qwen \citep{bai2023qwenvl}, Llama 4 Maverick \citep{meta2025llama4}, a Nemotron vision-language variant \citep{nvidia2024nemotron}, GPT \citep{openai2024gpt4}, and GLM-5.2 \citep{zhipu2024glm} are prompted zero-shot with a shared forensic verdict schema requesting a structured judgment (status, whether the image appears AI-manipulated, and a probability estimate) rather than a bespoke prompt engineered per model. Results were collected out-of-band, some via vendor API and some via browser automation, and imported into the same canonical schema by per-model adapters; this track therefore evaluates each model's zero-shot forensic reasoning ability rather than a tuned, purpose-built detection deployment.

\smallskip
\noindent\textbf{Open-source detectors (24 of 27 surveyed).} Twenty-four independently distributed open-source detectors are run locally on GPU against the full corpus, spanning standard Hugging Face \texttt{transformers} classifiers, custom Hugging Face loaders with bundled inference code, and academic checkpoints paired with their original repository's inference script. Seven of the twenty-four correspond to published academic methods: DRCT \citep{chen2024drct}, RINE \citep{koutlis2024rine}, C2P-CLIP \citep{tan2024c2pclip}, DeCLIP \citep{bitdefender2025declip}, AIDE \citep{meet2025aide}, GenD-DINOv3-L \citep{yermandy2026gend}, and the Community Forensics ViT-384 checkpoint \citep{owenslab2024commfor,hf_commfor_owenslab}. The remaining seventeen are individually distributed Hugging Face Hub community checkpoints without an associated paper: \texttt{ntire2026\_deepfake} \citep{hf_ntire2026}, \texttt{dima806\_ai\_vs\_real} \citep{hf_dima806}, \texttt{bombek1\_siglip\_dinov2} \citep{hf_bombek1}, \texttt{organika\_sdxl} \citep{hf_organika}, \texttt{aidfr\_real\_v2} \citep{hf_aidfr}, \texttt{nahrawy\_aiornot} \citep{hf_nahrawy}, \texttt{ummmaybe\_vit} \citep{hf_ummmaybe}, \texttt{yaya\_source} \citep{hf_yaya}, \texttt{ash\_flux\_vit} \citep{hf_ashflux}, \texttt{king1oo1\_deepguard} \citep{hf_king1oo1}, \texttt{haywoodsloan\_deploy} \citep{hf_haywoodsloan}, \texttt{date3k2\_vit} \citep{hf_date3k2}, \texttt{jacob\_distilled} \citep{hf_jacob_distilled}, \texttt{opensight\_commfor} \citep{hf_opensight_commfor}, \texttt{ateeqq\_siglip2} \citep{hf_ateeqq}, \texttt{sadra\_sdxl\_face} \citep{hf_sadra_sdxl_face}, and \texttt{wvolf\_vit} \citep{hf_wvolf}, the last of which originates from an MSc thesis at Solent University rather than a purely anonymous upload. For every community checkpoint we cite the exact Hugging Face repository path and uploader username as recorded in the project's own model registry; we do not independently re-verify the uploader's identity, training data, or claimed performance beyond what that registry and the checkpoint's own model card state. Each detector's native label convention is mapped onto the shared $P(\text{fake})$ scale according to its own documented semantics; label mappings are never inverted post hoc to inflate a score, so a detector that genuinely disagrees with this corpus's notion of ``fake'' surfaces a below-chance ROC-AUC rather than having that disagreement silently corrected away. Three additional, published detectors from the original 27-model survey, UniversalFakeDetect \citep{ojha2023ufd}, FatFormer \citep{liu2024fatformer}, and NPR \citep{tan2024npr}, were not integrated in this study and are excluded from the reported results; we treat this as a limitation (Section~\ref{sec:limitations}) rather than omit it silently.

\smallskip
\noindent\textbf{Fairness controls applied uniformly across all three tracks.} All 36 models score the identical 100-image corpus under the identical neutral-filename protocol described in Section~\ref{sec:dataset}. Failed or refused responses are marked as unsuccessful, excluded entirely from Accuracy, F1, ROC-AUC, and MCC, and surfaced separately as coverage, so that a model is neither rewarded nor penalized for declining to answer, but its abstention rate remains visible. Latency is compared only within the commercial track, since open-source inference time depends on local hardware and batching, and the imported LLM results were not uniformly timed; comparing latency across these heterogeneous execution environments would not be a fair comparison, so we do not attempt it.

\smallskip
\noindent\textbf{Score normalization for commercial adapters.} Because the five commercial vendors report native scores on entirely different scales and conventions, each is mapped to $P(\text{fake})$ according to its own documented output schema rather than a shared assumption. Hive, Reality Defender, and Sightengine each expose a normalized probability directly, which is used as $P(\text{fake})$ without transformation. TruthScan reports a numeric score on a documented 0--100 scale together with a categorical \texttt{final\_result} field; we take $P(\text{fake}) = \text{score}/100$ and use this same value, thresholded at 0.5, for both the hard label and ROC-AUC, so that a single probability drives every metric for this vendor, consistent with the protocol in Section~\ref{sec:metrics}. Neural Defend returns, per sub-check, a predicted label together with that check's confidence in its own prediction; we reconstruct $P(\text{fake})$ for each sub-check as its confidence when the check predicts fake and one minus its confidence when the check predicts genuine, then take the maximum across checks, so that the resulting score reflects the model's belief the image is fake rather than its belief in whichever label it happened to assign.

\subsection{Metrics}
\label{sec:metrics}

Let $TP$, $FP$, $TN$, and $FN$ denote true positives, false positives, true negatives, and false negatives, where the positive class is fake. For a model that produces $d$ successful verdicts out of the full corpus, all threshold-dependent metrics below are computed over those $d$ decided images only, at the standard 0.5 decision threshold on $P(\text{fake})$.

\begin{equation}
\text{Accuracy} = \frac{TP + TN}{d}
\label{eq:accuracy}
\end{equation}
Accuracy is the fraction of decided images classified correctly; on this corpus's 79:21 skew it is easy to game and, as we discuss below, should never be read in isolation.

\begin{equation}
\text{Precision} = \frac{TP}{TP + FP}
\label{eq:precision}
\end{equation}
Precision measures, among images flagged fake, what fraction genuinely are fake.

\begin{equation}
\text{Recall} = \frac{TP}{TP + FN}
\label{eq:recall}
\end{equation}
Recall (sensitivity, or true positive rate) measures the fraction of genuine fakes a model catches.

\begin{equation}
\text{Specificity} = \frac{TN}{TN + FP}
\label{eq:specificity}
\end{equation}
Specificity (the true negative rate) measures the fraction of the 21 authentic images correctly left alone, and is the direct measure of false-alarm behavior on this corpus's real-image controls.

\begin{equation}
F_1 = \frac{2 \cdot \text{Precision} \cdot \text{Recall}}{\text{Precision} + \text{Recall}}
\label{eq:f1}
\end{equation}
$F_1$ is the harmonic mean of precision and recall; like accuracy, it can be dominated by performance on the larger (fake) class alone.

\begin{equation}
\text{MCC} = \frac{TP \cdot TN - FP \cdot FN}{\sqrt{(TP+FP)(TP+FN)(TN+FP)(TN+FN)}}
\label{eq:mcc}
\end{equation}
The Matthews correlation coefficient ranges from $-1$ (total disagreement) to $+1$ (perfect prediction), with $0$ equivalent to chance once class prevalence is accounted for. Unlike accuracy or $F_1$, MCC only rises when a model performs well on \emph{both} classes simultaneously \citep{chico2020mcc}, which is why it is our \textbf{primary ranking metric}: a degenerate ``always predict fake'' classifier scores approximately $0.79$ accuracy and a deceptively high $F_1$ on this corpus, yet an MCC of approximately zero, correctly reflecting its complete lack of real discriminative skill.

ROC-AUC, our \textbf{tiebreak metric}, is computed from the ranked confidence scores rather than the thresholded hard labels, and is therefore threshold-free: it equals the probability that a randomly chosen fake image receives a higher $P(\text{fake})$ score than a randomly chosen real image, with $0.5$ corresponding to chance-level ranking and $1.0$ to perfect separation. MCC and ROC-AUC measure genuinely different things, operating-point quality at a fixed threshold versus pure ranking ability independent of any threshold, and while Section~\ref{sec:discussion} shows the two are strongly correlated across most of the leaderboard, a specific subset of models remain miscalibrated at their shipped threshold despite strong ROC-AUC, a one-directional risk that is itself a central empirical finding of this paper. Coverage is the fraction of the corpus for which a model returned a successful verdict at all. We do not perform pairwise significance testing across all $\binom{36}{2} = 630$ model comparisons in this study; we discuss this omission explicitly in Section~\ref{sec:limitations}.

\FloatBarrier
\section{Results}
\label{sec:results}

\begin{figure*}[!ht]
    \centering
    \includegraphics[width=\textwidth]{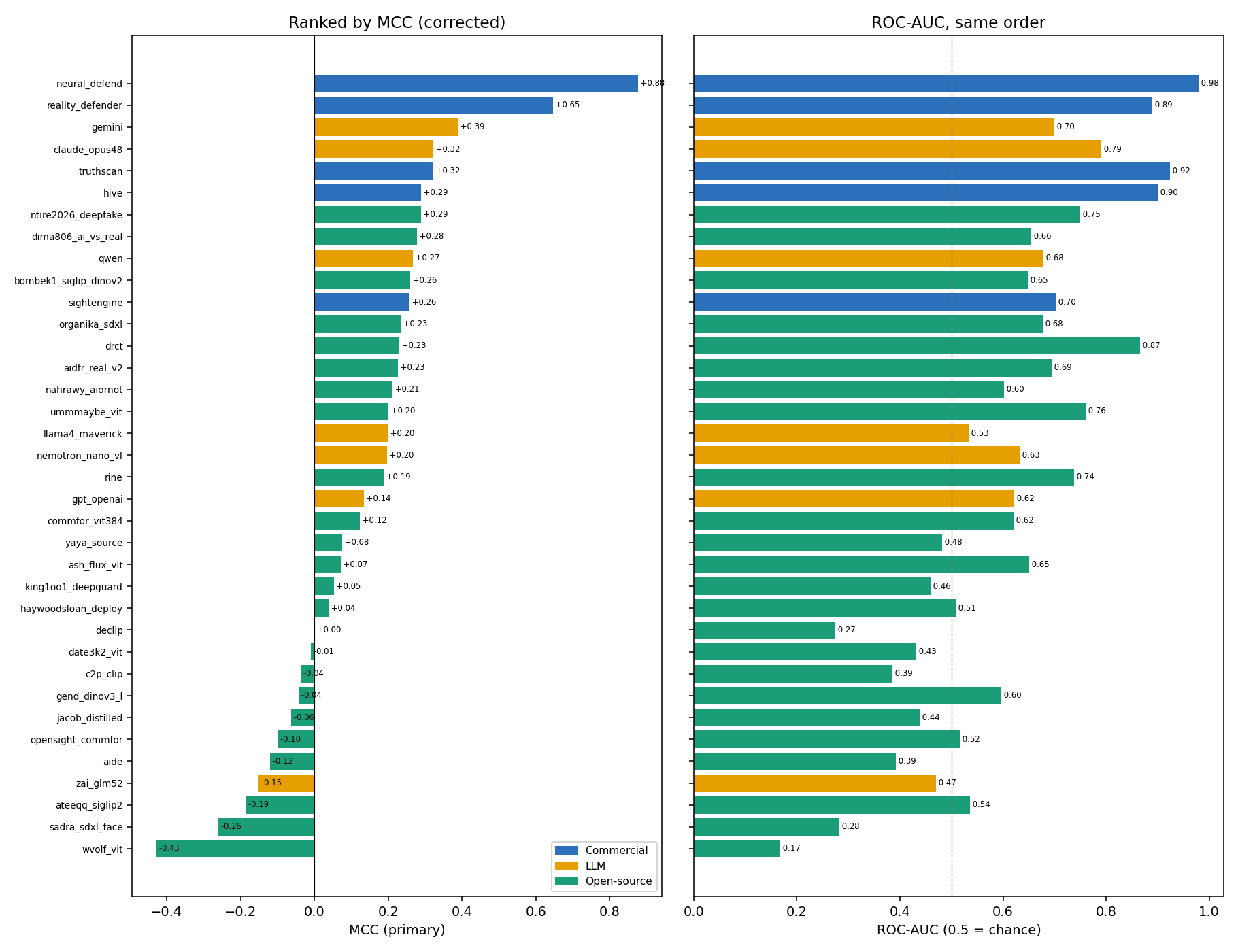}
    \caption{Unified leaderboard of all 36 models, ranked by MCC (primary) with ROC-AUC alongside. Neural Defend leads on both metrics; the ROC-AUC panel exposes the starved subset whose ranking power outstrips their operating-point quality.}
    \label{fig:leaderboard_img}
\end{figure*}

All 36 models were scored on the identical 100-image corpus (79 fake, 21 real) under the protocol of Section~\ref{sec:methodology}, ranked by MCC with ROC-AUC as the tiebreak criterion. ROC-AUC is reported alongside every threshold-dependent metric so that both ranking quality and operating-point performance can be interpreted from the same row. Table~\ref{tab:leaderboard} presents the complete unified leaderboard, combining all three evaluation tracks into a single ranking rather than separating commercial APIs, vision LLMs, and open-source detectors, so that the overall distribution of performance, where approaches cluster, and where they diverge, is visible directly rather than only the strongest performer within each category. Figure~\ref{fig:leaderboard_img} visualizes the complete ranking presented in Table~\ref{tab:leaderboard}. It highlights not only the ordering by MCC but also the accompanying ROC-AUC values, making it immediately apparent that several models exhibit strong disagreement between operating-point performance and ranking ability.

\begin{table*}[!t]
\centering
\caption{Unified leaderboard of all 36 evaluated models across the three benchmark tracks. Models are ranked by MCC (primary) with ROC-AUC used as the tie-break criterion. Positive class = fake.}
\label{tab:leaderboard}
\scriptsize
\setlength{\tabcolsep}{3pt}
\renewcommand{\arraystretch}{1.05}
\resizebox{\textwidth}{!}{%
\begin{tabular}{@{}rllrrrrrrrr@{}}
\toprule
\textbf{\#} &
\textbf{Model} &
\textbf{Track} &
\textbf{MCC} &
\textbf{AUC} &
\textbf{Acc.} &
\textbf{F1} &
\textbf{Prec.} &
\textbf{Rec.} &
\textbf{Spec.} &
\textbf{Cov.} \\
\midrule
1 & \texttt{neural\_defend}~\citep{neuraldefend2024docs} & Commercial & 0.876 & 0.979 & 0.960 & 0.975 & 0.963 & 0.987 & 0.857 & 0.99 \\
2 & \texttt{reality\_defender}~\citep{realitydefender2024realapi} & Commercial & 0.646 & 0.890 & 0.850 & 0.898 & 0.971 & 0.835 & 0.905 & 1.00 \\
3 & \texttt{gemini}~\citep{google2024gemini} & LLM & 0.389 & 0.700 & 0.610 & 0.678 & 0.976 & 0.519 & 0.952 & 1.00 \\
4 & \texttt{truthscan}~\citep{truthscan2024imageapi} & Commercial & 0.322 & 0.923 & 0.490 & 0.523 & 1.000 & 0.354 & 1.000 & 1.00 \\
5 & \texttt{claude\_opus48}~\citep{anthropic2024claudeopus} & LLM & 0.322 & 0.791 & 0.490 & 0.523 & 1.000 & 0.354 & 1.000 & 1.00 \\
6 & \texttt{hive}~\citep{hive2024genai} & Commercial & 0.290 & 0.900 & 0.450 & 0.466 & 1.000 & 0.304 & 1.000 & 1.00 \\
7 & \texttt{ntire2026\_deepfake}~\citep{hf_ntire2026} & Open-source & 0.289 & 0.750 & 0.600 & 0.683 & 0.915 & 0.544 & 0.810 & 1.00 \\
8 & \texttt{dima806\_ai\_vs\_real}~\citep{hf_dima806} & Open-source & 0.277 & 0.655 & 0.810 & 0.893 & 0.806 & 1.000 & 0.095 & 1.00 \\
9 & \texttt{qwen}~\citep{bai2023qwenvl} & LLM & 0.267 & 0.678 & 0.470 & 0.505 & 0.964 & 0.342 & 0.952 & 1.00 \\
10 & \texttt{bombek1\_siglip\_dinov2}~\citep{hf_bombek1} & Open-source & 0.259 & 0.649 & 0.570 & 0.650 & 0.909 & 0.506 & 0.810 & 1.00 \\
11 & \texttt{sightengine}~\citep{sightengine2024genai} & Commercial & 0.258 & 0.702 & 0.410 & 0.404 & 1.000 & 0.253 & 1.000 & 1.00 \\
12 & \texttt{organika\_sdxl}~\citep{hf_organika} & Open-source & 0.233 & 0.677 & 0.510 & 0.574 & 0.917 & 0.418 & 0.857 & 1.00 \\
13 & \texttt{drct}~\citep{chen2024drct} & Open-source & 0.230 & 0.866 & 0.470 & 0.514 & 0.933 & 0.354 & 0.905 & 1.00 \\
14 & \texttt{aidfr\_real\_v2}~\citep{hf_aidfr} & Open-source & 0.227 & 0.694 & 0.670 & 0.769 & 0.859 & 0.696 & 0.571 & 1.00 \\
15 & \texttt{nahrawy\_aiornot}~\citep{hf_nahrawy} & Open-source & 0.211 & 0.602 & 0.520 & 0.593 & 0.897 & 0.443 & 0.810 & 1.00 \\
16 & \texttt{ummmaybe\_vit}~\citep{hf_ummmaybe} & Open-source & 0.201 & 0.761 & 0.510 & 0.581 & 0.895 & 0.430 & 0.810 & 1.00 \\
17 & \texttt{llama4\_maverick}~\citep{meta2025llama4} & LLM & 0.199 & 0.533 & 0.340 & 0.283 & 1.000 & 0.165 & 1.000 & 1.00 \\
18 & \texttt{nemotron\_nano\_vl}~\citep{nvidia2024nemotron} & LLM & 0.196 & 0.632 & 0.394 & 0.374 & 0.944 & 0.233 & 0.952 & 0.94 \\
19 & \texttt{rine}~\citep{koutlis2024rine} & Open-source & 0.187 & 0.737 & 0.380 & 0.367 & 0.947 & 0.228 & 0.952 & 1.00 \\
20 & \texttt{gpt\_openai}~\citep{openai2024gpt4} & LLM & 0.135 & 0.621 & 0.370 & 0.364 & 0.900 & 0.228 & 0.905 & 1.00 \\
21 & \texttt{commfor\_vit384}~\citep{owenslab2024commfor} & Open-source & 0.123 & 0.620 & 0.550 & 0.651 & 0.840 & 0.532 & 0.619 & 1.00 \\
22 & \texttt{yaya\_source}~\citep{hf_yaya} & Open-source & 0.075 & 0.482 & 0.730 & 0.836 & 0.802 & 0.873 & 0.190 & 1.00 \\
23 & \texttt{ash\_flux\_vit}~\citep{hf_ashflux} & Open-source & 0.072 & 0.650 & 0.500 & 0.597 & 0.822 & 0.468 & 0.619 & 1.00 \\
24 & \texttt{king1oo1\_deepguard}~\citep{hf_king1oo1} & Open-source & 0.053 & 0.459 & 0.510 & 0.614 & 0.812 & 0.494 & 0.571 & 1.00 \\
25 & \texttt{haywoodsloan\_deploy}~\citep{hf_haywoodsloan} & Open-source & 0.038 & 0.508 & 0.410 & 0.468 & 0.812 & 0.329 & 0.714 & 1.00 \\
26 & \texttt{declip}~\citep{bitdefender2025declip} & Open-source & 0.000 & 0.275 & 0.210 & 0.000 & 0.000 & 0.000 & 1.000 & 1.00 \\
27 & \texttt{date3k2\_vit}~\citep{hf_date3k2} & Open-source & $-$0.010 & 0.432 & 0.340 & 0.353 & 0.783 & 0.228 & 0.762 & 1.00 \\
28 & \texttt{c2p\_clip}~\citep{tan2024c2pclip} & Open-source & $-$0.038 & 0.385 & 0.370 & 0.422 & 0.767 & 0.291 & 0.667 & 1.00 \\
29 & \texttt{gend\_dinov3\_l}~\citep{yermandy2026gend} & Open-source & $-$0.043 & 0.597 & 0.340 & 0.365 & 0.760 & 0.241 & 0.714 & 1.00 \\
30 & \texttt{jacob\_distilled}~\citep{hf_jacob_distilled} & Open-source & $-$0.063 & 0.439 & 0.400 & 0.483 & 0.757 & 0.354 & 0.571 & 1.00 \\
31 & \texttt{opensight\_commfor}~\citep{hf_opensight_commfor} & Open-source & $-$0.100 & 0.517 & 0.530 & 0.667 & 0.758 & 0.595 & 0.286 & 1.00 \\
32 & \texttt{aide}~\citep{meet2025aide} & Open-source & $-$0.119 & 0.392 & 0.400 & 0.500 & 0.732 & 0.380 & 0.476 & 1.00 \\
33 & \texttt{zai\_glm52}~\citep{zhipu2024glm} & LLM & $-$0.152 & 0.470 & 0.710 & 0.830 & 0.772 & 0.899 & 0.000 & 1.00 \\
34 & \texttt{ateeqq\_siglip2}~\citep{hf_ateeqq} & Open-source & $-$0.187 & 0.536 & 0.620 & 0.762 & 0.753 & 0.772 & 0.048 & 1.00 \\
35 & \texttt{sadra\_sdxl\_face}~\citep{hf_sadra_sdxl_face} & Open-source & $-$0.260 & 0.282 & 0.250 & 0.257 & 0.591 & 0.165 & 0.571 & 1.00 \\
36 & \texttt{wvolf\_vit}~\citep{hf_wvolf} & Open-source & $-$0.428 & 0.168 & 0.220 & 0.278 & 0.517 & 0.190 & 0.333 & 1.00 \\
\bottomrule
\end{tabular}%
}
\end{table*}

\subsection{Overview of the Unified Leaderboard}
\label{sec:leaderboard-overview}

Table~\ref{tab:leaderboard} makes several qualitative patterns visible before any per-track breakdown is needed. The top of the table is dominated by the two commercial APIs that decide every image confidently and consistently, and the gap between this leading pair and the rest of the field is the widest single gap anywhere in the ranking, suggesting these two systems are doing something structurally different from every other model rather than merely performing incrementally better. Immediately below them, the field becomes a genuine mixture of all three paradigms rather than a clean paradigm-by-paradigm block: strong vision LLMs, the best open-source rankers, and the remaining commercial APIs interleave closely, which is itself informative, since it means paradigm membership alone does not reliably predict where a given model lands once the leading pair is set aside.

The bottom of the table tells a different and equally informative story. Every model with a negative or near-zero score is an open-source detector or a vision LLM; all five commercial APIs post a positive MCC, and several of the lower-ranked models achieve superficially reasonable accuracy or F1 figures despite a default decision rule that collapses toward predicting a single class almost regardless of the image it is shown, which is exactly the pathology the ranking metric is designed to expose rather than reward. Reading the table column by column rather than row by row reinforces this: models that look strong on accuracy or F1 do not always look strong on specificity or on ranking power, and the ordering implied by any single column frequently disagrees with the ordering implied by another. No single column of this table should be read as the leaderboard on its own; the ranking metric is deliberately chosen to resist exactly this kind of single-column overinterpretation, and Section~\ref{sec:discussion} unpacks the most consequential instances of that disagreement in detail.

\subsection{Per-Track Summary}

By operating-point quality, commercial APIs lead, vision LLMs sit mid-tier, and open-source detectors trail, consistent with limited exposure to this corpus's modern generators. 

\begin{figure}[!ht]
    \centering
    \includegraphics[width=\columnwidth]{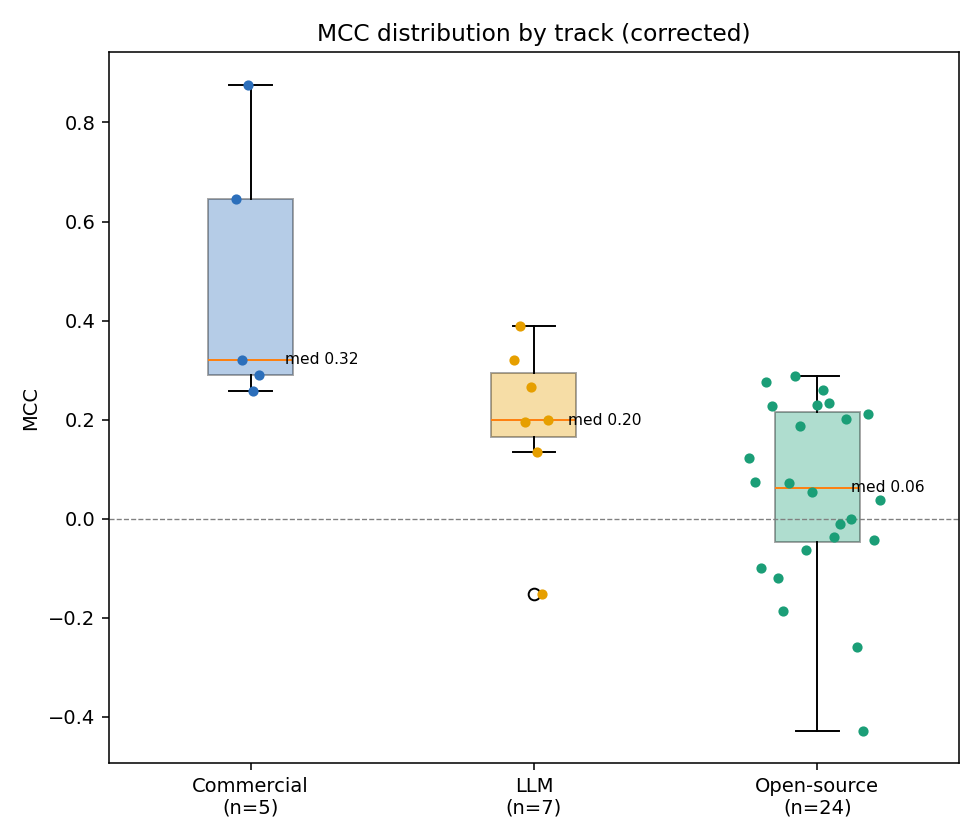}
    \caption{Distribution of MCC values across the three tracks. Commercial APIs show the highest median operating-point quality (0.32), while open-source detectors show the widest spread.}
    \label{fig:track_mcc}
\end{figure}

Ranking power complicates this picture only in the open-source track: the strongest vision LLM does not reach the ranking power of the single best open-source detector, even though the open-source track's typical performance is by far the weakest of the three; a single well-designed academic method can out-rank an entire tier of expensive, general-purpose foundation models on pure class separation, even while the median detector in that same open-source track performs far worse than either alternative paradigm. The commercial track retains the single strongest results on both metrics simultaneously, neural\_defend leads the entire study on MCC and on ROC-AUC, which is itself informative: it shows that the metric-reliability risk this paper documents in Section~\ref{sec:discussion} is a property of specific miscalibrated models, not an inherent feature of the paradigm they belong to. Figure~\ref{fig:track_mcc} summarizes the distribution of MCC values within each benchmark track. Commercial APIs exhibit the highest median operating-point quality, whereas the open-source ecosystem shows substantially larger variability, indicating that performance differences within that paradigm are considerably greater than between the commercial offerings. Table~\ref{tab:pertrack} aggregates the leaderboard into paradigm-level statistics, reporting the strongest and median performance within each evaluation track.

\begin{table}[htbp]
\centering
\caption{Per-track summary. Models = count evaluated; Coverage = mean across the track's models.}
\label{tab:pertrack}
\small
\begin{tabular}{@{}lrlrlrr@{}}
\toprule
\textbf{Track} & \textbf{Models} & \textbf{Best MCC} & \textbf{Median MCC} & \textbf{Best AUC} & \textbf{Median AUC} & \textbf{Mean Cov.} \\
\midrule
Commercial API & 5 & 0.876 (neural\_defend) & 0.322 & 0.979 (neural\_defend) & 0.900 & 100\% \\
Vision LLM & 7 & 0.389 (gemini) & 0.199 & 0.791 (claude\_opus48) & 0.632 & 99\% \\
Open-source & 24 & 0.289 (ntire2026) & 0.062 & 0.866 (drct) & 0.566 & 100\% \\
\bottomrule
\end{tabular}
\end{table}

\subsection{Latency (Commercial APIs Only)}

The five commercial providers span a response-time range wide enough to have direct deployment consequences independent of accuracy. The fastest provider responds quickly enough for genuinely interactive, real-time use, while the slowest is an order of magnitude too slow for any synchronous flow and is realistically suited only to asynchronous, batch-style pipelines where an extra delay per item is immaterial.

\begin{figure*}[!ht]
    \centering
    \includegraphics[width=\textwidth]{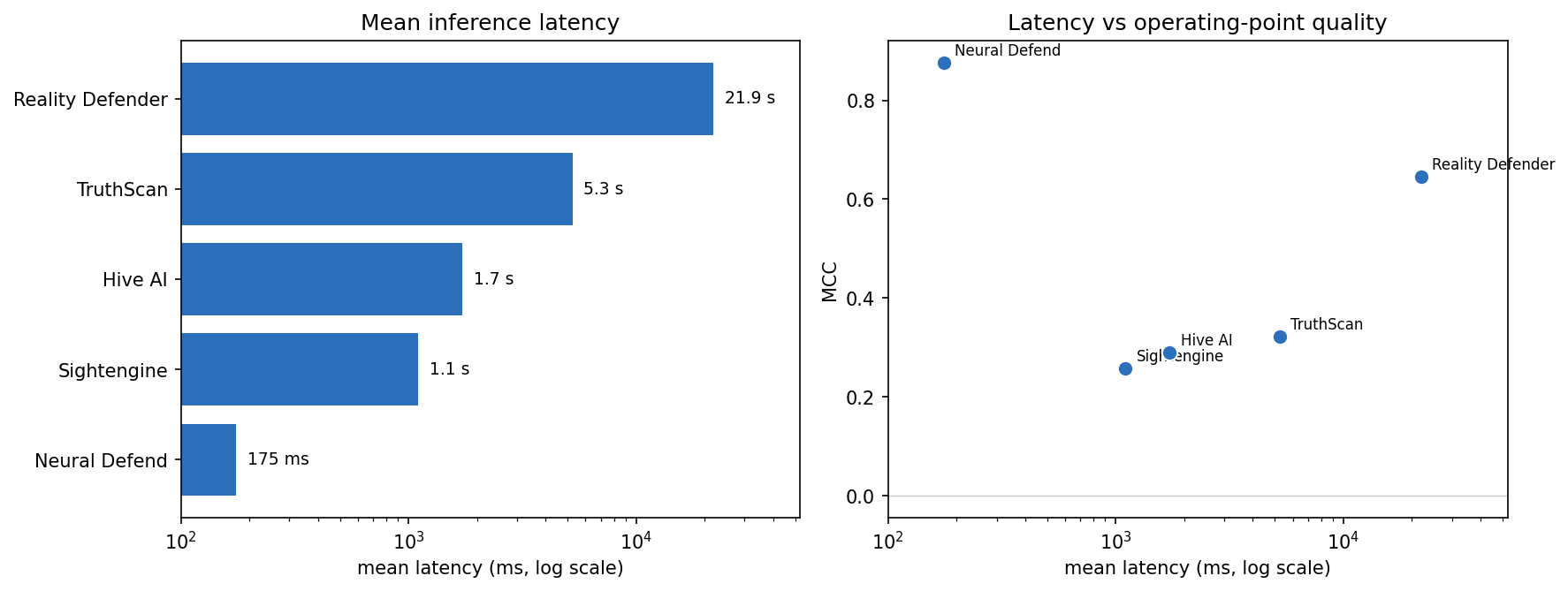}
    \caption{Latency comparison of the five commercial APIs. Left: mean inference latency (log scale). Right: latency versus MCC. The fastest provider (Neural Defend, 175 ms) also achieves the highest operating-point quality, while slower/costlier services do not reliably buy better detection.}
    \label{fig:latency}
\end{figure*}

 Latency and metric quality are not otherwise linked in any simple way: the fastest provider in the study also posts the strongest operating-point quality and the strongest ranking power simultaneously, while providers among the slowest span a wide range of both metrics, so a procurement decision cannot assume that a slower or more expensive service buys better detection. Figure~\ref{fig:latency} illustrating both the latency distribution of the commercial APIs and its relationship with operating-point quality. The figure shows that low inference latency does not necessarily require sacrificing detection performance, as the fastest provider also achieves the highest MCC. Table~\ref{tab:latency} summarizes the average inference latency of the five commercial APIs evaluated in this study. Since latency depends heavily on execution environment, comparisons are restricted to the commercial track, where all services were measured under a consistent evaluation setup.

\begin{table}[htbp]
\centering
\caption{Mean latency, commercial APIs only. Latency is not comparable across tracks (Section~\ref{sec:methodology}).}
\label{tab:latency}
\small
\begin{tabular}{@{}lr@{}}
\toprule
\textbf{Provider} & \textbf{Mean latency} \\
\midrule
Neural Defend & 175\,ms \\
Sightengine & 1.1\,s \\
Hive AI & 1.7\,s \\
TruthScan & 5.3\,s \\
Reality Defender & 21.9\,s \\
\bottomrule
\end{tabular}
\end{table}

\FloatBarrier
\section{Discussion}
\label{sec:discussion}

\begin{figure}[!ht]
    \centering
    \includegraphics[width=\columnwidth]{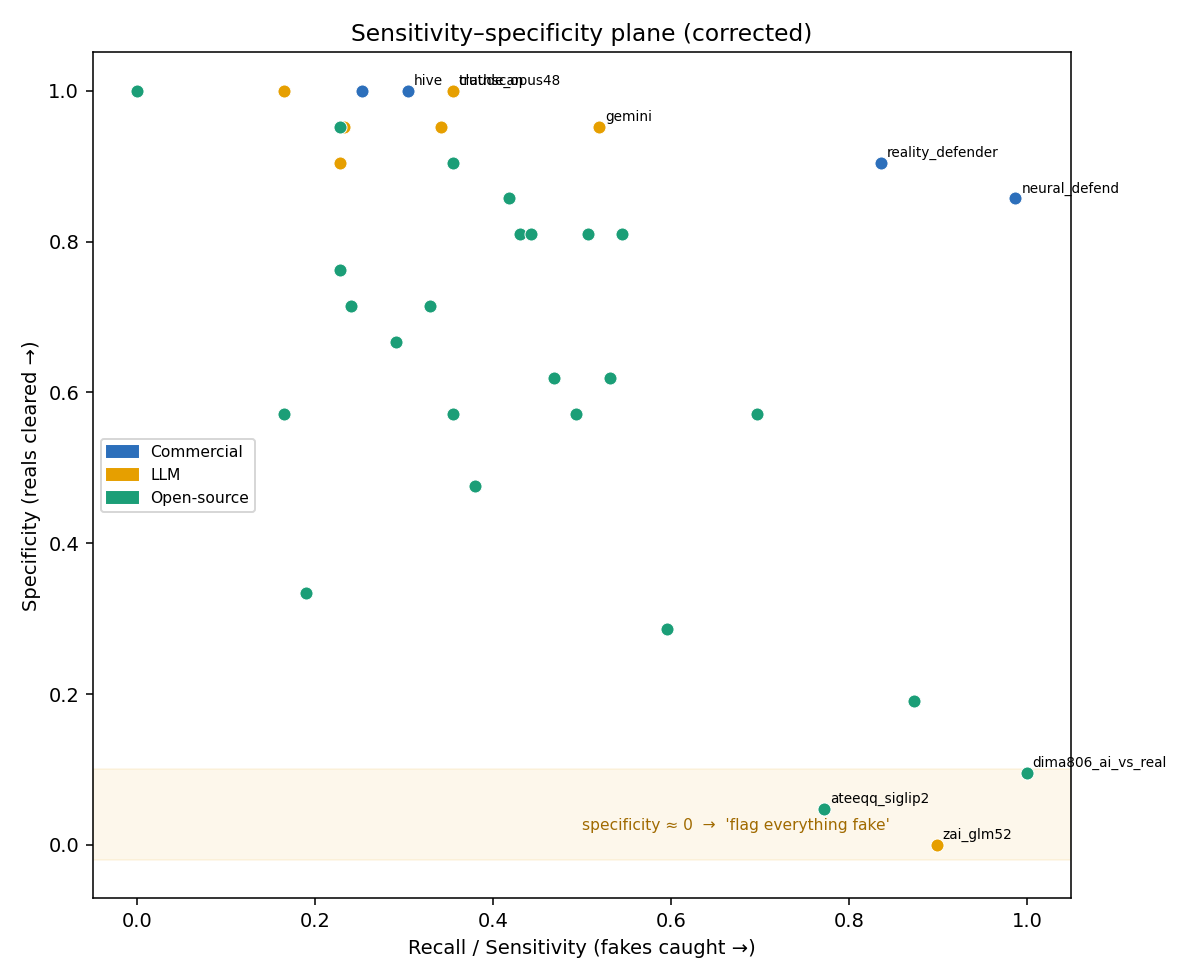}
    \caption{Sensitivity–specificity plane for all models. Neural Defend and Reality Defender occupy the high-recall/high-specificity corner; a base-rate-exploiting cluster collapses to specificity $\approx$ 0.}
    \label{fig:sensitivity_specificity}
\end{figure}

Figure~\ref{fig:sensitivity_specificity} illustrates the trade-off between sensitivity and specificity across all evaluated detectors. Several models cluster near perfect recall but very low specificity, revealing a strong tendency to classify nearly every image as fake, whereas the strongest detectors maintain a substantially better balance between detecting manipulations and avoiding false alarms. This distribution reinforces that high recall alone is insufficient for deployment, since practical forensic systems must identify synthetic media without overwhelming users with false alarms on genuine content.

The plane divides the field into three practically distinct populations rather than a single continuum. A small cluster in the upper-right, occupied by Neural Defend and Reality Defender, achieves both high recall and high specificity simultaneously; these are the only two models in the entire study that a deployment could plausibly trust without a compensating human-review step, since they rarely miss a fake and rarely flag a genuine photograph. A second, much larger cluster sits along the top edge at high specificity but with recall spread widely from near-zero to moderate; this is where most open-source detectors, several vision LLMs, and TruthScan all land, reflecting models that are conservative by default rather than genuinely discriminating, a pattern consistent with the recall-versus-specificity trade-off already visible in the per-model metrics of Table~\ref{tab:leaderboard}. A third population collapses toward the bottom-right corner, combining high recall with specificity near zero; \texttt{dima806\_ai\_vs\_real}, \texttt{ateeqq\_siglip2}, and \texttt{zai\_glm52} occupy this region, and each achieves its recall not through genuine discrimination but by defaulting toward ``fake'' on nearly every input, the same base-rate exploitation this paper returns to below. No model in the study occupies the bottom-left corner, meaning nothing evaluated here is simultaneously bad at catching fakes and bad at preserving real images; every detector fails, when it fails, in one of these two specific, interpretable directions rather than randomly.

\paragraph{A Clear Paradigm Ordering, With a Genuine Exception}

The headline ordering commercial APIs ahead of vision LLMs ahead of the open-source median is unsurprising on its face: commercial vendors have direct financial incentive to optimize detection accuracy specifically, vision LLMs are prompted zero-shot for a task they were never trained on, and most publicly distributed open-source detectors were trained against generator families that predate the modern, adversarial mix this corpus targets. What is not unsurprising is that this ordering is a paradigm-level statement about typical performance, not a universal one. DRCT, a single open-source ranker built on contrastive reconstruction training \citep{chen2024drct}, reaches ROC-AUC 0.866, ahead of every vision LLM in the study including the strongest, Claude Opus 4.8, at 0.791. A practitioner who dismissed the entire open-source track on the basis of its poor median performance would have missed the one detector in that track genuinely competitive with, and by this metric better than, general-purpose multimodal LLMs costing far more per query to run.

\begin{figure*}[!ht]
    \centering
    \includegraphics[width=\textwidth]{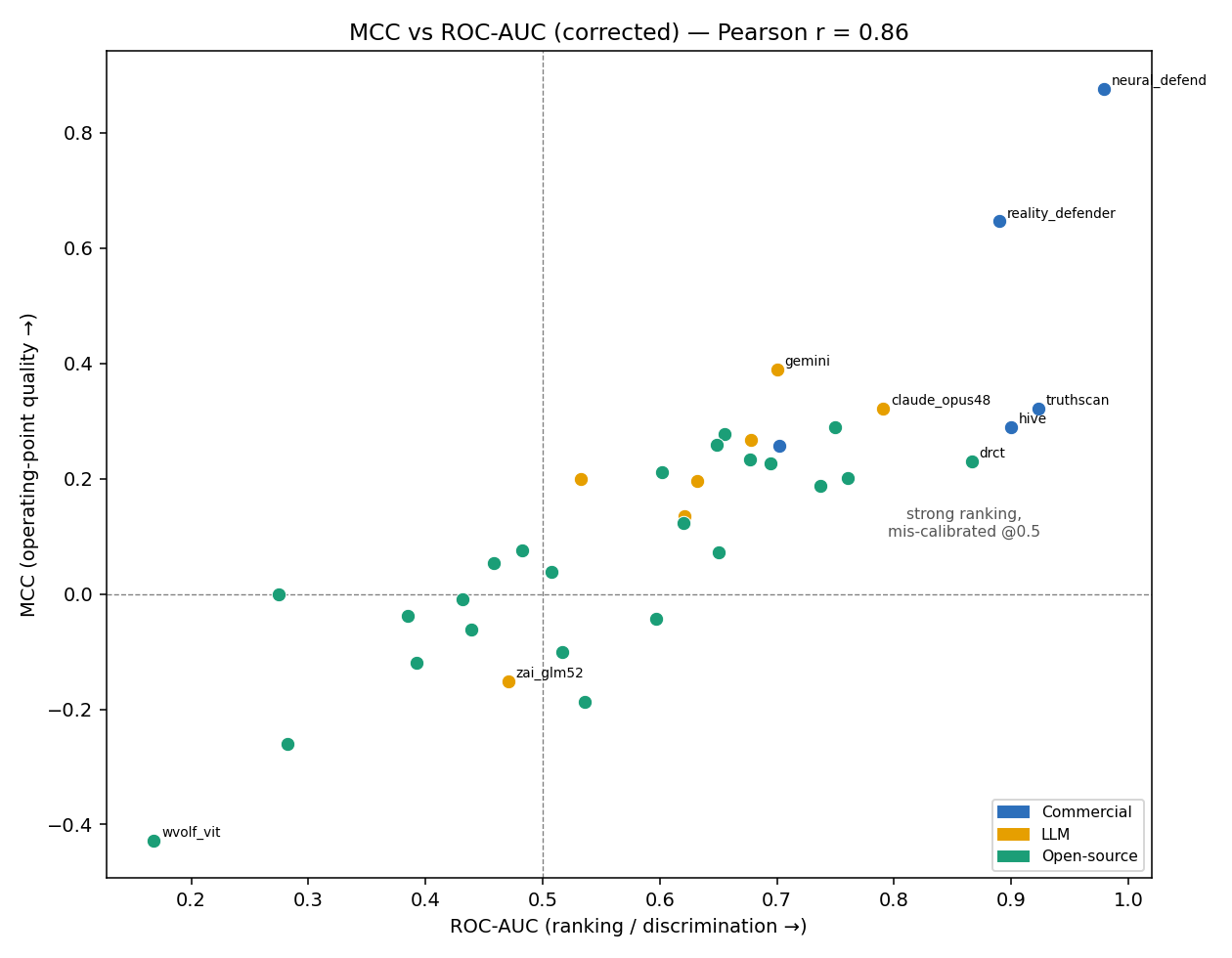}
    \caption{Relationship between operating-point quality (MCC) and ranking ability (ROC-AUC). Most of the 36 models fall close to the diagonal where the two metrics agree (Pearson r $\approx$ 0.86); a starved subset sits below it; threshold miscalibration, not weak discrimination.}
    \label{fig:mcc_auc}
\end{figure*}

\paragraph{AUC Can Overstate Deployability}

Ranking power and operating-point quality are not opposing forces in this study. Across all 36 models, MCC and ROC-AUC are strongly correlated (Pearson $r \approx 0.86$, Spearman $\approx 0.87$): the model with the best MCC, Neural Defend, is also the model with the best ROC-AUC (0.979), and the model with the weakest MCC, \texttt{wvolf\_vit}, is also among the weakest on ROC-AUC. For most of the field, a strong ranker is also a well-calibrated one, and vice versa. The finding worth reporting is narrower, and one-directional, rather than a bidirectional paradox: a subset of models separate the two classes well by ROC-AUC yet under-deliver on MCC at their shipped 0.5 threshold, so ROC-AUC alone would overstate how deployable they actually are. TruthScan, Hive, DRCT, and Claude Opus 4.8 are the clearest examples, each posting a ROC-AUC above 0.79 while their MCC sits meaningfully below what that ranking power might suggest (0.322, 0.290, 0.230, and 0.322 respectively). We never observe the reverse pattern at any comparable magnitude, no model in the leaderboard combines strong MCC with weak ROC-AUC, which is precisely why the finding is one-directional: a high ROC-AUC is necessary but not sufficient evidence of a trustworthy default decision, while a strong MCC is rarely accompanied by a surprisingly weak ROC-AUC.

Figure~\ref{fig:mcc_auc} makes this concrete. The bulk of the field lies close to the diagonal running from the origin toward the upper-right corner, where MCC and ROC-AUC rise together, which is the ordinary case this adversarial corpus is designed to make less automatic than it would be on an easier benchmark. A starved band sits visibly below that diagonal: TruthScan, Hive, DRCT, and Claude Opus 4.8 all reach ROC-AUC above 0.79 while their MCC lags behind, the signature of a model whose confidence scores separate the classes correctly in principle but whose shipped decision threshold does not translate that separation into the best achievable hard-label agreement with ground truth. No comparable band appears above the diagonal, reinforcing that this is a one-directional risk specific to threshold calibration, not a general unpredictability between the two metrics.

\begin{figure*}[!ht]
    \centering
    \includegraphics[width=\textwidth]{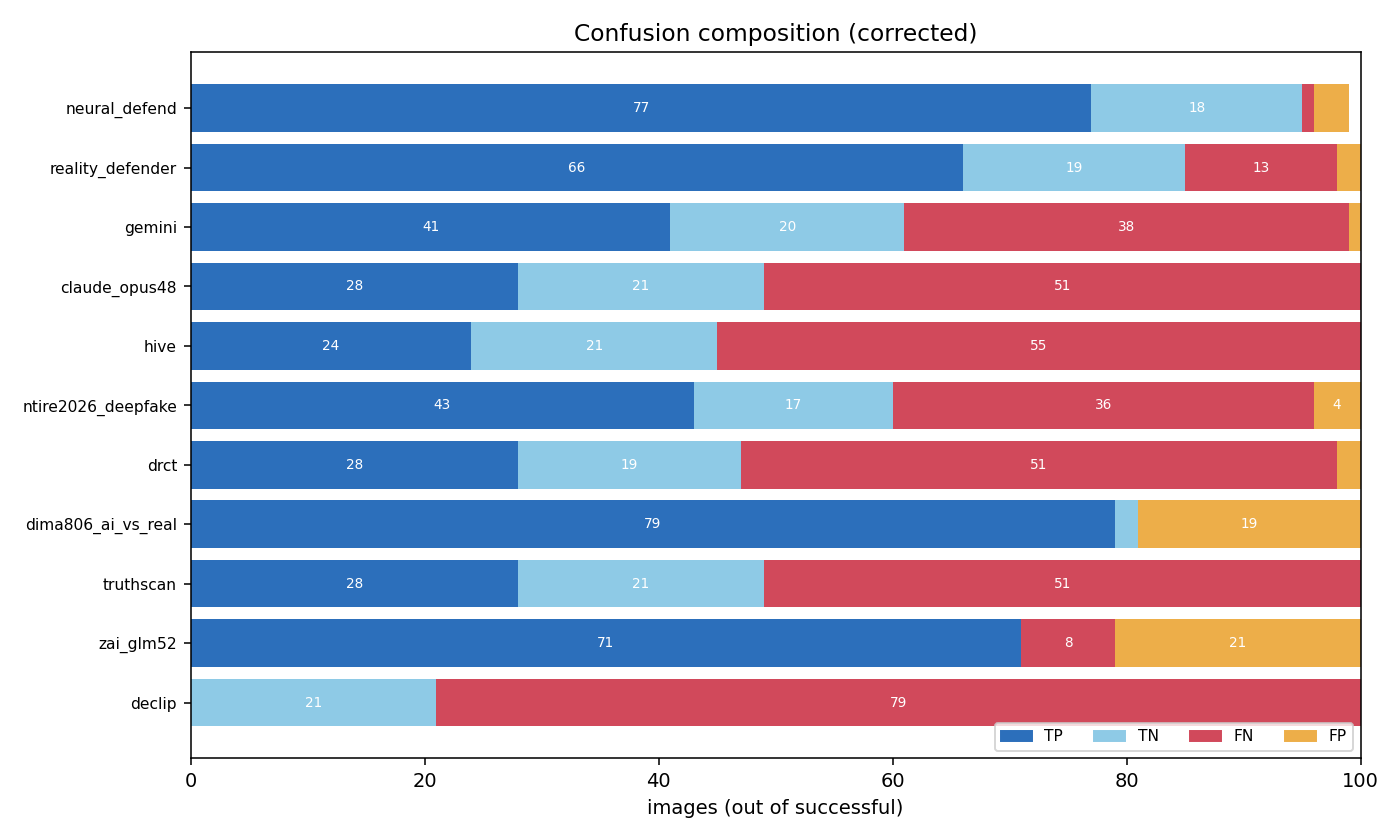}
    \caption{Confusion composition for representative detectors. Each stacked bar splits the decided images into TP/TN/FN/FP, showing why similar-accuracy models get very different MCC.}
    \label{fig:confusion}
\end{figure*}

Figure~\ref{fig:confusion} shows why: models with superficially similar accuracy produce markedly different confusion compositions. Neural Defend and Reality Defender are dominated by true positives with only a thin false-positive sliver; \texttt{dima806\_ai\_vs\_real}, \texttt{ateeqq\_siglip2}, and \texttt{zai\_glm52} show almost no true-negative bar at all, the base-rate exploitation this paper returns to below; and DRCT's false-negative bar is nearly as large as TruthScan's or Claude Opus 4.8's, the visual cost of the same conservative default threshold that suppresses its MCC despite ranking fakes above reals better than any other open-source detector in the study. TruthScan and Claude Opus 4.8 in fact share an identical confusion composition, a full true-negative bar and a substantial false-negative bar, the signature of two models from different paradigms converging on the same conservative operating point. No single bar or metric predicts MCC in isolation; it is the balance across all four confusion cells that the ranking metric is built to capture.

\paragraph{The Base-Rate Trap Is Not Hypothetical}

Because the corpus is 79\% fake by design, a model that predicts fake indiscriminately scores deceptively well on accuracy and F1 while carrying zero real discriminative skill. This is not a theoretical caveat: \texttt{zai\_glm52} is the concrete instance among the vision LLMs, posting 71.0\% accuracy and F1 of 0.830 while its specificity is exactly zero (it flags all 21 real images as fake) and its MCC is negative ($-0.152$). \texttt{dima806\_ai\_vs\_real} and \texttt{ateeqq\_siglip2} exhibit the identical pathology in the open-source track, with specificity of 0.095 and 0.048 respectively despite accuracy figures (0.810 and 0.620) that would look unremarkable in isolation. Any evaluation of these systems that reported accuracy or F1 without also reporting specificity and MCC would have rated all three as reasonable performers; none is.

\paragraph{Scope of These Claims}

Every result in this paper is a point measurement on one fixed, deliberately adversarial, 100-image corpus, evaluated in a single run per model. We do not compute confidence intervals or perform pairwise significance testing across the $\binom{36}{2} = 630$ possible model comparisons in this study, both because the corpus is too small to support many of the resulting per-comparison tests meaningfully and because a systematic multiple-comparisons correction across 630 tests was out of scope for this study; we return to this explicitly as future work in Section~\ref{sec:limitations}. The vision-LLM results in particular were collected out-of-band rather than through a uniformly controlled live harness, so that track's numbers should be read as a generic zero-shot baseline rather than a tuned, purpose-built deployment. We interpret every number in this paper as a diagnostic signal on a hard, curated stress test, not as an estimate of population-level accuracy in deployment.

\FloatBarrier
\section{Limitations and Future Work}
\label{sec:limitations}

Several limitations bound the conclusions of this study, each paired with the extension it motivates. First, the corpus is small (100 images) and imbalanced (79:21), limiting confidence in individual metrics, while the smallest per-generator groups, several with only one or two images, are too small for stable rate estimates; a corpus of several hundred to a few thousand images, scaled up proportionally across the same eight edge-case families rather than diluted by easy, homogeneous content, would tighten every reported estimate and let per-generator-group breakdowns support statistically meaningful comparisons rather than anecdote. Second, every result reflects a single run per model, without multiple seeds, bootstrap resampling, confidence intervals, or significance testing across the 630 possible pairwise comparisons; adding bootstrap confidence intervals, multi-seed runs, and a principled subset of statistical comparisons is a direct extension of this work. Third, the vision-LLM results were collected out-of-band, some via vendor APIs and others through browser automation, rather than through the uniformly controlled live harness used for the commercial track, making that track less reproducible. Fourth, provenance for 6 of the 21 fake source groups remains partial or unverified, and three open-source detectors from the original 27-model survey were not integrated in this study. Fifth, the score-normalization conventions described in Section~\ref{sec:normalization} were validated by cross-checking each vendor's documented output schema against a sample of raw responses; extending this validation to every response for every model, rather than a representative sample, is a natural and worthwhile hardening step for any future extension of this benchmark. Finally, the generator mix reflects a 2026 snapshot and will inevitably evolve as new generators and detectors emerge.
 
Future work includes growing the corpus from 100 to several hundred or a few thousand images while preserving its adversarial, edge-case-heavy design, adding new source groups and deeper coverage of the thinnest existing ones rather than bulk content that would let easy cases dilute the diagnostic signal; adding bootstrap confidence intervals and multi-seed runs for greater statistical rigor; completing the full 27-model open-source survey by resolving the three deferred detectors; producing per-generator-family analyses to identify where each paradigm fails; extending full-sample adapter validation to every response for every model as a standard, automated step before any result is reported; and investigating calibrated, per-model decision thresholds rather than shipped defaults, which Section~\ref{sec:discussion} shows can substantially change operating-point quality without affecting underlying discriminative ability.

\FloatBarrier
\section{Conclusion}
\label{sec:conclusion}
 
This paper set out to answer a question no single prior evaluation could: given a detection budget, does paying for a commercial API, prompting a general-purpose vision-language model already available, or self-hosting a free open-source detector actually buy meaningfully different capability against modern, adversarial synthetic media? We built VendorBench-100, a cross-paradigm benchmark evaluating 36 models including 5 commercial APIs, 7 zero-shot vision LLMs, and 24 open-source detectors on a single, deliberately hard 100-image corpus under one shared normalization schema, one abstention policy, and one ranking framework. The headline answer is that paradigm matters on average: commercial APIs achieve the strongest typical performance, vision LLMs occupy a middle tier despite never being trained for this task, and most open-source detectors trail on this adversarial generator mix. Yet this average conceals important exceptions, including DRCT, which outperforms every vision LLM on ranking ability, demonstrating that carefully designed academic detectors remain competitive despite the rapid evolution of proprietary detection systems. This finding carries a direct implication for research investment: sustained work on generalizable, open detection methods remains scientifically justified, not merely a lower-cost fallback.
 
The more consequential result is that, while MCC and ROC-AUC are strongly correlated across the 36 models (Pearson $r \approx 0.86$), so that for most detectors the two properties move together and reporting either one alone is usually a reasonable, if incomplete, summary, a specific, identifiable subset of models see their ROC-AUC substantially overstate their operating-point reliability, separating the classes well in principle while under-delivering at their shipped default threshold. This one-directional miscalibration risk appears across multiple paradigms and, together with the class-imbalance accuracy trap this benchmark is built to expose, demonstrates that no single scalar metric is safe to report in isolation on an imbalanced, adversarial corpus. For a practitioner, the actionable consequence is concrete: any procurement or deployment decision should weigh MCC, specificity, and ROC-AUC together rather than trust a single leaderboard score, since any one of them in isolation can make a genuinely strong detector look weak, or a genuinely unreliable one look strong, depending on which axis is omitted. By releasing the complete evaluation harness, per-image evidence, and aggregated results, we provide a reproducible foundation for future benchmark development and more rigorous cross-paradigm comparisons. We believe this evaluation philosophy will encourage future benchmarks to prioritize fair, standardized, and practically meaningful comparisons over isolated leaderboard performance alone.
 
Beyond the specific 36 models evaluated here, VendorBench-100 represents a shift toward benchmarking paradigm choice itself rather than individual detectors in isolation. As commercial services evolve behind proprietary interfaces, vision-language models continue to expand beyond their original purpose, and open-source checkpoints proliferate rapidly, meaningful evaluation must remain unified, reproducible, and continuously updated. We hope the adversarial corpus design, shared normalization protocol, and complementary use of MCC and ROC-AUC become standard practice for future deepfake benchmarks, while larger corpora, repeated evaluations, statistical significance testing, and video-based extensions provide an increasingly reliable basis for selecting trustworthy detection systems.


\bibliographystyle{unsrt}
\bibliography{references,references_extra}

\end{document}